\documentclass[sigconf]{acmart}
\AtBeginDocument{%
  }

\setcopyright{acmlicensed}
\copyrightyear{2026}
\acmYear{2026}
\setcopyright{cc}
\setcctype{by}
\acmConference[KDD 2026] {Proceedings of the 32nd ACM SIGKDD Conference on Knowledge Discovery and Data Mining V.2}{August 9--13, 2026}{Jeju Island, Republic of Korea.}
\acmBooktitle{Proceedings of the 32nd ACM SIGKDD Conference on Knowledge Discovery and Data Mining V.2 (KDD 2026), August 9--13, 2026, Jeju Island, Republic of Korea}
\acmDOI{10.1145/3770855.3817473}
\acmISBN{979-8-4007-2259-2/2026/08}

\usepackage{booktabs}
\usepackage{nicefrac}       
\usepackage{microtype}      

\usepackage{graphicx}
\usepackage{multirow}

\usepackage{lineno}
\usepackage{subfigure}
\usepackage{fontawesome}
\usepackage{mathtools}

\usepackage{hyperref}
\usepackage{url}
\usepackage{algorithm}
\usepackage{amsfonts}
\usepackage{algorithmic}
\usepackage{amsmath}
\usepackage{booktabs}
\usepackage{newfloat}
\usepackage{listings}
\usepackage{xcolor}
\usepackage{tcolorbox}
\usepackage{colortbl}
\usepackage{enumitem}
\usepackage{wrapfig}
\usepackage{etoc}

\definecolor{backred}{RGB}{255, 190, 190}
\definecolor{backblue}{RGB}{220, 230, 250}
\definecolor{myblue}{RGB}{0, 102, 204}

\newcommand{\blueurl}[1]{\textcolor{myblue}{\url{#1}}}

\begin{document}

\title{Speak-to-Structure: Evaluating LLMs in Open-domain Natural Language-Driven Molecule Generation}


\author{Jiatong Li}
\authornote{Both authors contributed equally to this research.}
\email{jiatong.li@connect.polyu.hk}
\orcid{0000-0001-7705-2296}
\affiliation{%
  \institution{Hong Kong Polytechnic University}
  \country{Hong Kong SAR}
}

\author{Junxian Li}
\authornotemark[1]
\email{lijunxian0531@sjtu.edu.cn}
\affiliation{%
  \institution{Shanghai Jiao Tong University}
  \city{Shanghai}
  \country{China}
  }

\author{Weida Wang}
\email{wangweida@pjlab.org.cn}
\affiliation{%
  \institution{Shanghai AI Lab}
  \city{Shanghai}
  \country{China}
}

\author{Yunqing Liu}
\email{yunqing617.liu@connect.polyu.hk}
\affiliation{%
  \institution{Hong Kong Polytechnic University}
  \country{Hong Kong SAR}
 }

\author{Changmeng Zheng}
\email{changmeng.zheng@polyu.edu.hk}
\affiliation{%
  \institution{Hong Kong Polytechnic University}
  \country{Hong Kong SAR}
  }

\author{Yatao Bian}
\email{ybian@nus.edu.sg}
\affiliation{%
  \institution{National University of Singapore}
  \country{Singapore}
}

\author{Dongzhan Zhou}
\authornote{Corresponding authors}
\email{zhoudongzhan@pjlab.org.cn}
\affiliation{%
  \institution{Shanghai AI Lab}
  \city{Shanghai}
  \country{China}
  }

\author{Xiao-Yong Wei}
\authornotemark[2]
\email{x1wei@polyu.edu.hk}
\affiliation{%
  \institution{Hong Kong Polytechnic University}
  \country{Hong Kong SAR}
  }

\author{Qing Li}
\email{qing-prof.li@polyu.edu.hk}
\affiliation{%
  \institution{Hong Kong Polytechnic University}
  \country{Hong Kong SAR}
  }

\renewcommand{\shortauthors}{Li et al.}

\begin{abstract}
Recently, Large Language Models (LLMs) have demonstrated great potential in natural language-driven molecule discovery. 
However, existing datasets and benchmarks for molecule-text alignment are predominantly built on one-to-one mappings, measuring LLMs' ability to retrieve a single, pre-defined answer, rather than their creative potential to generate diverse, yet equally valid, molecular candidates.
To address this critical gap, we propose \textbf{S}peak-to-\textbf{S}tructure (\textbf{S\textsuperscript{2}-Bench}), 
the first benchmark to evaluate LLMs in open-domain natural language-driven molecule generation.
S\textsuperscript{2}-Bench is specifically designed for one-to-many relationships, challenging LLMs to exhibit genuine molecular understanding and open-ended generation capabilities. 
Our benchmark includes three key tasks: molecule editing (\textbf{MolEdit}), molecule optimization (\textbf{MolOpt}), and customized molecule generation (\textbf{MolCustom}), each probing a different aspect of molecule discovery. 
We also introduce \textbf{OpenMolIns}, a large-scale instruction tuning dataset that enables Llama3.1-8B to surpass the most powerful LLMs like GPT-4o and Claude-3.5 on S\textsuperscript{2}-Bench. 
Our comprehensive evaluation of 31 LLMs shifts the focus from simple pattern recall to realistic molecular design, paving the way for more capable LLMs in natural language-driven molecule discovery.
Our codes and datasets are fully accessible through the \textbf{Github Repository}: \blueurl{https://github.com/phenixace/S2-TOMG-Bench} and \textbf{Huggingface Datasets}: \blueurl{https://huggingface.co/datasets/phenixace/S2-TOMG-Bench}.
\end{abstract}




\maketitle

\section{Introduction}
\label{sec:intro}

Molecule discovery plays a pivotal role in various scientific research fields \citep{liu2026enhancing,liu2025glprotein,zhang2025graphatc,cao2022identifying,li2026molvibench}, from pharmaceuticals \citep{keiser2010chemical} to materials science \citep{higuchi2023material}. Traditionally, this is a trial-and-error process \citep{ekins2024lab} that requires extensive experimentation and data analysis \citep{mattern2023automated}, often taking over a decade to bring a new drug candidate to market \citep{lee2018evaluating}.

Recently, Large Language Models (LLMs) have demonstrated great potential in molecule discovery \citep{edwards2021text2mol, edwards2022translation, chitsaz2025novomolgen, wang2025chem,li2025chemvlm} by leveraging their powerful language understanding and robust reasoning abilities \citep{achiam2023gpt}. Molecules can be represented as linearized representations like SMILES \citep{weininger1988smiles} and SELFIES \cite{krenn2022selfies}, which allows LLMs to process molecules as textual strings, effectively bridging the gap between molecular structures and natural language. Meanwhile, by aligning molecules with textual descriptions \citep{edwards2022translation,chen2025hierarchical}, LLMs can assist chemists in interpreting chemical knowledge, suggesting structural modifications, and predicting compound properties \citep{li2024large, zhang2024chemllm}, thereby significantly streamlining the molecule discovery process.

While the integration of LLMs holds immense promise, a significant challenge lies in the limitations of current datasets and benchmarks. For example, existing datasets for aligning molecules and texts like ChEBI-20 \citep{edwards2022translation}, PubChem324K \citep{liu2023molca}, and M$^3$-20M \citep{guo2025m3}, are constructed based on the one-to-one mapping assumption, where each textual description is linked to a single, predefined target molecule due to the convenience of acquiring ground truth data from existing databases. However, the one-to-one mapping presents a key limitation: it does not fully align with the nature of real-world molecule discovery. In practice, multiple distinct molecules can often share the same desired properties or biological activity \citep{petrone2012rethinking}. For instance, a particular pharmaceutical effect is rarely unique to a single compound, and a material with a specific physical property, like tensile strength, can be realized through various molecular structures. This mismatch between the evaluation paradigm and the reality of molecule discovery casts doubt on whether these datasets and benchmarks genuinely assess the capacity of LLMs for creative molecular design, or if they instead inadvertently encourage models to rely on memorization and pattern-matching.

\begin{figure*}
    \centering
    \includegraphics[width=0.95\linewidth]{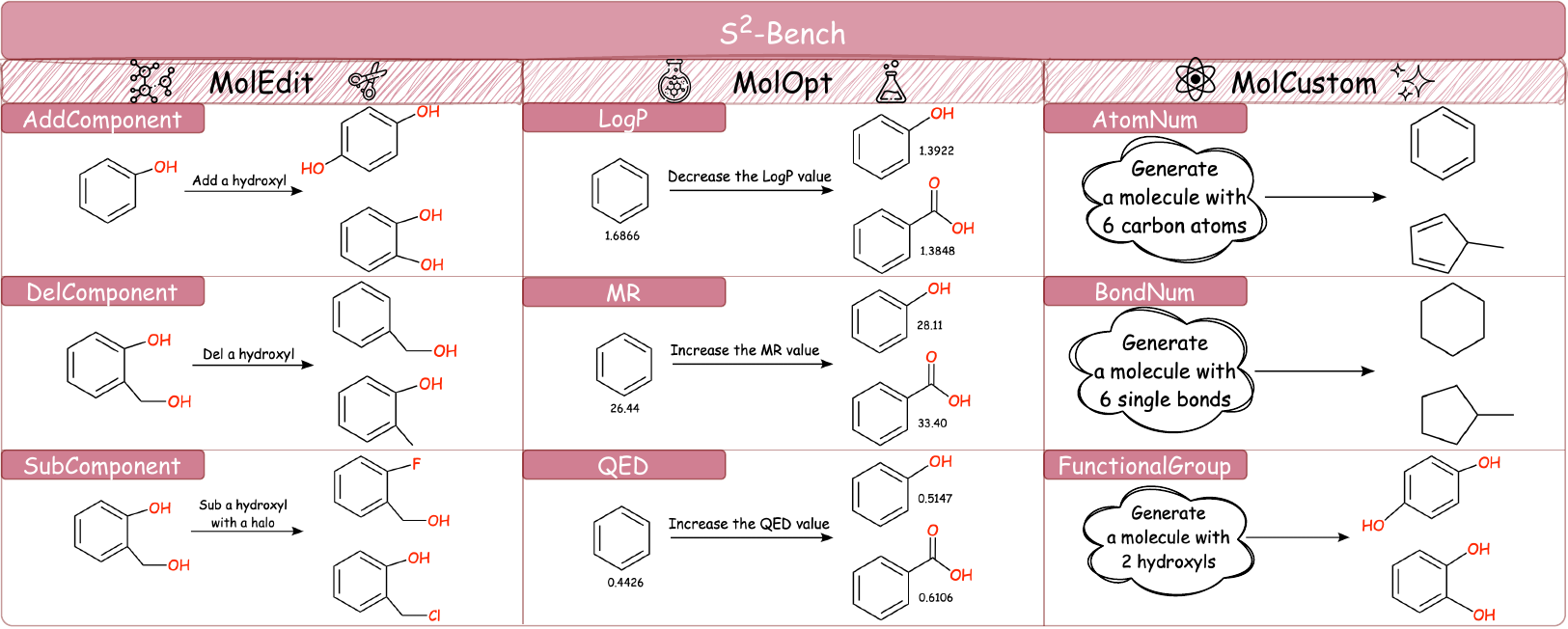}
    \caption{Task illustration of S\textsuperscript{2}-Bench for open domain natural language-driven molecule generation. In contrast to text-based target molecule generation, multiple valid molecules may fulfill the textual requirements (right of the arrow).}
    \Description{}
    \label{fig:tasks}
\end{figure*}

To bridge this critical gap, we introduce a novel benchmark, \textbf{S}peak-to-\textbf{S}tructure (\textbf{S\textsuperscript{2}-Bench}), the first benchmark designed to evaluate the open-domain natural language-driven molecule generation capabilities of LLMs. 
S\textsuperscript{2}-Bench is well structured into three primary tasks, as shown in Figure \ref{fig:tasks}, each targeting a specific, real-world capability essential for drug and materials science:

\noindent\textbf{Molecule Editing (MolEdit)} focuses on testing an LLM's ability to perform precise, localized structural modifications while preserving main structures. In drug discovery, this is analogous to Lead Optimization \citep{barcelos2022lead}, which edits the lead compound to create new variants. This requires a deep understanding of chemical grammar, including valency, stereochemistry, and ring systems, ensuring that edits could create valid molecules.

\noindent\textbf{Molecule Optimization (MolOpt)} extends the MolEdit task by evaluating a model’s ability to optimize a lead compound under specified property constraints (e.g., increasing solubility or reducing toxicity). Its open-ended design requires the model to edit and refine molecular structures in accordance with the desired properties, thereby showcasing its understanding of both structural modifications and property-related characteristics.

\noindent\textbf{Customized Molecule Generation (MolCustom)}  requires LLMs to synthesize a novel molecule based on a natural language description of its desired structural components. Specifically, it challenges models to generate molecules with a set of quantitative and qualitative constraints, such as a specific number of atoms, bonds, or functional groups, which directly mimics the initial drug design phase where a chemist defines a new compound based on a set of precise requirements for its structure. 

To facilitate the evaluation of the above tasks, we introduce an automated evaluation system tailored to open-domain molecule generation, where exact string matching is neither appropriate nor sufficient.
Specifically, we use cheminformatics toolkits to (i) verify \emph{validity} by parsing the generated outputs; (ii) check \emph{constraint satisfaction} (e.g., functional-group edits for MolEdit, property direction for MolOpt, and atom/bond/functional-group constraints for MolCustom); and (iii) quantify \emph{quality} via task-specific metrics, including structural similarity to the source molecule for MolEdit/MolOpt and novelty against a reference database for MolCustom.
This design enables a scalable and objective assessment of whether a model follows instructions, performs rational local edits, and generates chemically meaningful candidates.

Furthermore, we propose \textbf{OpenMolIns}, a large-scale instruction-tuning dataset comprising up to 120 million instruction-molecule pairs constructed programmatically from large chemical databases.
OpenMolIns mirrors the three tasks in S\textsuperscript{2}-Bench and provides diverse prompt templates to improve robustness to phrasing variations.

To summarize, our contributions are primarily threefold:

\begin{itemize}[itemsep=2pt,topsep=0pt,parsep=0pt]
\item[1.] We introduce \textbf{S}peak-to-\textbf{S}tructure (\textbf{S\textsuperscript{2}-Bench}), the first benchmark for open-domain natural language-driven molecule generation. Moving from one-to-one to one-to-many relationships, S\textsuperscript{2}-Bench better aligns concepts in drug discovery and provides a novel perspective on assessing LLMs' genuine molecular understanding and design capabilities.

\item[2.] We introduce \textbf{OpenMolIns}, an instruction-tuning dataset with up to 1.2 million instruction–molecule pairs, enabling Llama3.1-8B to achieve SOTA performance on S\textsuperscript{2}-Bench.

\item[3.] We provide insightful findings based on our extensive benchmarking of 31 LLMs, revealing the limitations of existing targeted generation datasets and highlighting the potential for LLMs to transition from simple pattern recall to realistic molecular design.
\end{itemize}
\section{Related Work}
\label{sec:relatedwork}

Recent breakthroughs in natural language processing (NLP) have highlighted the potential of Large Language Models (LLMs) to analyze complex biological and chemical data more effectively than traditional computational methods~\citep{zhou2023uni}. Within this context, Text-based Molecule Generation has emerged as a representative task for AI-driven molecule discovery~\citep{edwards2021text2mol}. This task focuses on generating target molecules directly from natural language descriptions, which requires the construction of paired datasets linking molecular structures with their textual representations.

Early work in this direction employed transformer-based models such as MolT5~\citep{edwards2022translation}, which leveraged large-scale self-supervised learning to produce high-quality SMILES strings from textual prompts. Building on this foundation, models like KV-PLM~\citep{zeng2022deep}, MoMu~\citep{su2022molecular}, and BioT5~\citep{pei2023biot5} integrated molecular graphs and biochemical texts to jointly enhance molecular understanding and generation. 

Parallel to these architecture-driven improvements, LLMs such as MolReGPT~\citep{li2024empowering} and ICMA~\citep{li2024large} have demonstrated strong in-context learning capabilities, enabling adaptive molecule generation by retrieving and utilizing relevant examples from the provided context. Besides, MolReFlect~\citep{li2024molreflect}, emphasizes fine-grained alignment between molecular structures and natural language descriptions through a teacher–student paradigm, which effectively captures subtle structural–semantic relationships. Moreover, Mol-R1 \citep{li2025mol} takes a step further by introducing DeepSeek-R1 like Long-CoT reasoning in molecule generation, enabling robust and interpretable natural language-driven molecule generation.

Meanwhile, \citet{nicolaou2013multi} propose the task of multi-objective molecular optimization and \citet{wu2024leveraging} further extend it to textual requirements by using prompt engineering to guide LLM-based molecule optimization.
In addition, \citet{ye2025drugassist} incorporate explicit structural constraints into the optimization process, while InstructMol \citep{zhuang2025advancing} further proposes an LLM that can understand and design biomolecules following human instructions.
\section{Speak-to-Structure (S\textsuperscript{2}-Bench)}
\label{sec:methodlogy}

In this section, we detail the design philosophy, technical composition, and statistics of Speak-to-Structure (\textbf{S\textsuperscript{2}-Bench}), which is fundamentally structured around the core capabilities required for real-world molecular design: editing, optimizing, and customized generation.

\subsection{Task Composition}
S\textsuperscript{2}-Bench is meticulously designed to move beyond simple pattern-matching and evaluate the genuine molecular understanding and generation capabilities of LLMs. We adopt a one-to-many paradigm, where a single language instruction can be fulfilled by multiple valid molecules. Our benchmark challenges LLMs to demonstrate a flexible understanding of molecular syntax rather than mere memorization. This design philosophy is concretely realized through three tasks of increasing complexity, each mirroring a critical phase of molecule discovery.\\

\noindent \textbf{MolEdit} 
assesses an LLM's foundational molecular syntax and structural modification capabilities. In this task, a base molecule is provided, and the model is asked to perform a specific change on the molecular structure. In a real-world scenario, this task is analogous to the iterative lead optimization process in drug discovery, where chemists make small, precise changes to a candidate molecule to change its properties. We designed three subtasks to probe this capability: \textit{AddComponent} and \textit{DelComponent} test the model's ability to add or remove a specific functional group, respectively, while \textit{SubComponent} challenges a combined operation of both.\\

\noindent\textbf{MolOpt} 
evaluates an LLM's ability to perform goal-oriented chemical reasoning. The challenge here is twofold: the LLM should not only edit a molecule but also ensure that the modification leads to a desired change in a specific property. This task directly mirrors the hit-to-lead optimization stage, where molecules are refined to achieve better pharmacological profiles. To assess this, we choose three key properties (\textit{LogP}, \textit{MR}, and \textit{QED}), which are vital for drug discovery and can be \textbf{reliably and reproducibly calculated}. By requiring the model to generate a molecule that improves on a given property, we test its ability to understand the complex relationship between molecular structure and function, a capability that cannot be evaluated by tasks with a single, predetermined answer.\\

\noindent\textbf{MolCustom} 
serves as the ultimate test of an LLM's creative chemical design ability. MolCustom requires the LLM to generate a novel molecule from a natural language description of its desired structural components. This task is not about modifying an existing structure but about synthesizing a new one from scratch, akin to a chemist defining the needs for a new compound at the very beginning of a project. To ensure a clear and precise evaluation of this complex task, we focus on generating molecules with specific structural features: a defined number of atoms (\textit{AtomNum}), a specified number and type of bonds (\textit{BondNum}), or the inclusion of particular functional groups (\textit{FunctionalGroup}). While these constraints may seem simple, they are deceptively challenging, requiring the model to apply a sophisticated understanding of chemical valence rules to create a valid and novel molecule.

Compared to targeted molecule generation datasets, these tasks in S\textsuperscript{2}-Bench, while seemingly straightforward in their instructions, impose a more rigorous requirement on LLMs' ability to perform precise, open-domain molecule generation. Our benchmark fundamentally shifts the evaluation from pattern recall to practical chemical design, thereby providing a more accurate measure of an LLM's potential in molecule discovery and helping to build more explainable and trustworthy models.

\subsection{Data Construction}
Previously, the development of robust datasets for text-based targeted molecule generation has been hindered by the scarcity of high-quality human annotations. For example, while text-to-image generation datasets like MS COCO \citep{chen2015microsoft} contain millions of annotated samples, molecule-caption datasets, such as ChEBI-20 \citep{edwards2022translation} are significantly smaller, often by orders of magnitude. This disparity arises because molecular annotation demands specialized expertise, making it both time-intensive and costly. The resulting data scarcity further poses a significant challenge for advancing LLMs in text-guided molecule discovery.

Our approach to data construction fundamentally bypasses this bottleneck. S\textsuperscript{2}-Bench is designed for open-domain molecule generation tasks that do not rely on human annotations. Instead, we leverage automated chemical toolkits to programmatically construct tasks and evaluate outputs based on objective molecular properties and structural rules. This enables us to generate a virtually unlimited volume of data, effectively addressing the ``data hunger" that has constrained previous research. More importantly, this programmatic approach allows us to create tasks with a one-to-many relationship, where multiple correct answers exist for a single prompt, which better aligns with the complexities of drug discovery, accommodating its inherent variability and diverse outcomes.

Specifically, we design a systematic and scalable data construction process, as shown in Figure \ref{fig:data}.

For \textbf{MolEdit} and \textbf{MolOpt}, we sample molecules from large, publicly available chemical databases. We selected Zinc-250K \citep{sterling2015zinc} for building our test set due to its manageable size and diversity, while the massive PubChem \citep{kim2019pubchem} database (with over 10 million molecules) serves as the basis for the instruction-tuning dataset. We utilize the RDKit \citep{landrum2013RDKit} toolbox to automatically extract key molecular statistics, including structural patterns and properties like LogP, MR, and QED. These extracted features are then integrated into our pre-defined, instruction-based prompt templates.

For \textbf{MolCustom}, we move beyond existing databases and generate instructions to test a model's ability to create novel molecules. For each of the three subtasks, we randomly generate 5,000 instructions that specify a target number and type of atoms, bonds, or functional groups.

Furthermore, for each subtask, we pre-defined a diverse prompt template pool to ensure that the LLMs are not overfit to a limited set of prompt formats. This programmatic construction not only allows for the generation of a much larger data volume but also ensures that our benchmark tests an LLM's intrinsic chemical reasoning ability, rather than its capacity for memorizing human-annotated patterns. By moving beyond the one-to-one paradigm, S\textsuperscript{2}-Bench provides a more robust and realistic evaluation, paving the way for models that can truly innovate and design novel molecules.

\subsection{Evaluation}
The evaluation of S\textsuperscript{2}-Bench is facilitated through a set of carefully designed automated evaluation processes and metrics tailored to the unique nature of our tasks. Unlike one-to-one datasets or benchmarks that simply check for exact matches, our metrics are designed to assess an LLM's ability to produce valid, relevant, and novel outputs within an open-ended framework.

\begin{table}[htbp]
    \centering
    \caption{All used metrics are listed here. '\checkmark' means that a metric is calculated on a task.}
    \resizebox{1.0\columnwidth}{!}{
    \begin{tabular}{lcccc}
    \toprule
        \rowcolor{gray!20} \textbf{Task} & \textbf{(Weighted) Success Rate} & \textbf{Similarity} &\textbf{Novelty} & 
        \textbf{Validity} \\
        \midrule 
        MolEdit & \checkmark & \checkmark& & \checkmark\\         
        MolOpt & \checkmark&\checkmark & & \checkmark\\
        MolCustom & \checkmark& & \checkmark& \checkmark\\
    \bottomrule
    \end{tabular}
    }
    
    \label{tab:metric}
\end{table}

\subsubsection{MolEdit \& MolOpt Evaluation}
For the \textbf{MolEdit} and \textbf{MolOpt} tasks, as shown in Table \ref{tab:metric}, we employ a combination of metrics to assess both the correctness of the modification and the rationality of the design.

\textbf{Success Rate}: This metric evaluates a model's ability to fulfill the specific molecule generation requirements, with values ranging from 0 to 1. We design automated evaluation processes to verify if the generated molecule meets the specified criteria (e.g., correct functional group modification for MolEdit, or desired property optimization for MolOpt). This metric examines whether the LLM can follow instructions precisely, which is a fundamental requirement for an LLM to serve as a chemist assistant. 

\textbf{Similarity}: In open-domain tasks like MolEdit and MolOpt, a high Success Rate alone is insufficient. We must also ensure that the generated molecule is a reasonable modification of the original, rather than a completely new, unrelated structure that happens to satisfy the criteria. To address this, we measure the Tanimoto Similarity between the generated and original molecules using Morgan Fingerprints \citep{butina1999unsupervised}. A high similarity score indicates that the model has performed a rational, localized edit and has not simply generated a different molecule from scratch. The similarity $\delta(m^g, m^o) \in [0,1]$ between the generated molecule $m^g$ and the original molecule $m^o$ is calculated as:
\begin{equation}
    \delta(m^g, m^o) = \frac{\left| fp_{m^g} \cap fp_{m^o} \right|}{\left| fp_{m^g} \cup fp_{m^o} \right|},
\end{equation}
where $fp_{m^g}$ and $fp_{m^o}$ represent their corresponding Morgan Fingerprints. $\left| fp_{m^g} \cap fp_{m^o} \right|$ is the size of the intersection between their Morgan Fingerprints, while $\left| fp_{m^g} \cup fp_{m^o} \right|$ is the union.
    
\textbf{Validity}: This metric evaluates the percentage of the generated molecules that are chemically valid and follow the rules of molecular syntax. Molecules need to successfully pass the SMILES parser, and a higher validity means that the model is more familiar with the molecule syntax.

\subsubsection{MolCustom Evaluation}
For the \textbf{MolCustom} task, where models are required to generate molecules from scratch based on structural descriptions, we use a different set of metrics to evaluate their creative design capabilities.

\noindent \textbf{Success Rate}: This metric measures how well the generated molecules adhere to the high-level structural constraints provided in the prompt (e.g., number of atoms, bonds, or specific functional groups). As the task requires de novo generation, the success rate directly reflects a model's ability to translate abstract requirements into a valid, concrete molecular structure. 

\noindent \textbf{Novelty}: For open-domain generation, fulfilling the requirements is just the first step. The true value lies in a model's ability to generate novel and innovative molecules. The novelty score quantifies this by comparing the generated molecules against existing structures in a large database, such as Zinc-250K (Zinc for short). A low similarity to known molecules suggests a high degree of novelty, which is a critical indicator of a model's potential for discovering truly new molecule structures.
The novelty $n$ for the generated molecule $m^g$ can be calculated as:
\begin{equation}
    n(m^g) = 1 - \max_{m^k \in \text{Zinc}} \delta(m^g, m^k)
\end{equation}
\noindent \textbf{Validity}: As with the other two tasks, validity ensures that all generated molecules are chemically sound and can be interpreted correctly.

\subsubsection{Average Weighted Success Rate}
To provide a single, comprehensive ranking of LLM performance on S\textsuperscript{2}-Bench, we introduce a weighted success rate. This metric combines the core success rate with a quality metric relevant to each task: Similarity for MolEdit/MolOpt and Novelty for MolCustom. This approach ensures that a high score reflects not only the ability to follow instructions but also the rationality and creativity of the generation. The weighted success rate for a subtask $t$ is defined as:
\begin{equation}
    W\!S\!R_t = \begin{cases}
        n_t\times S\!R_t,  & t \in \{MolCustom\} \\
        \delta_t\times S\!R_t, & t \in \{MolEdit, MolOpt\}
    \end{cases},
\end{equation}
where $W\!S\!R_t$ denotes the weighted success rate for a subtask $t$, while $\delta_t$ is the similarity score for the MolEdit and MolOpt tasks, $n_t$ represents the novelty score for the MolCustom tasks, and $S\!R_t$ is the corresponding success rate of the subtask. Then, the average weighted success rate $\overline{W\!S\!R}$ could evaluate the synthetic performance of LLMs among all the nine subtasks:
\begin{equation}
    \overline{W\!S\!R}=\frac{1}{9}\sum_{t}W\!S\!R_t,
\end{equation}
where the average weighted success rate $\overline{W\!S\!R}$ provides a balanced measure of a model's performance across all nine subtasks, offering a robust and nuanced assessment of its capabilities in open-domain natural language-driven molecule generation.

\subsection{OpenMolIns: Instruction Tuning Dataset}

\begin{figure*}[htbp]
    \centering
    \includegraphics[width=0.8\linewidth]{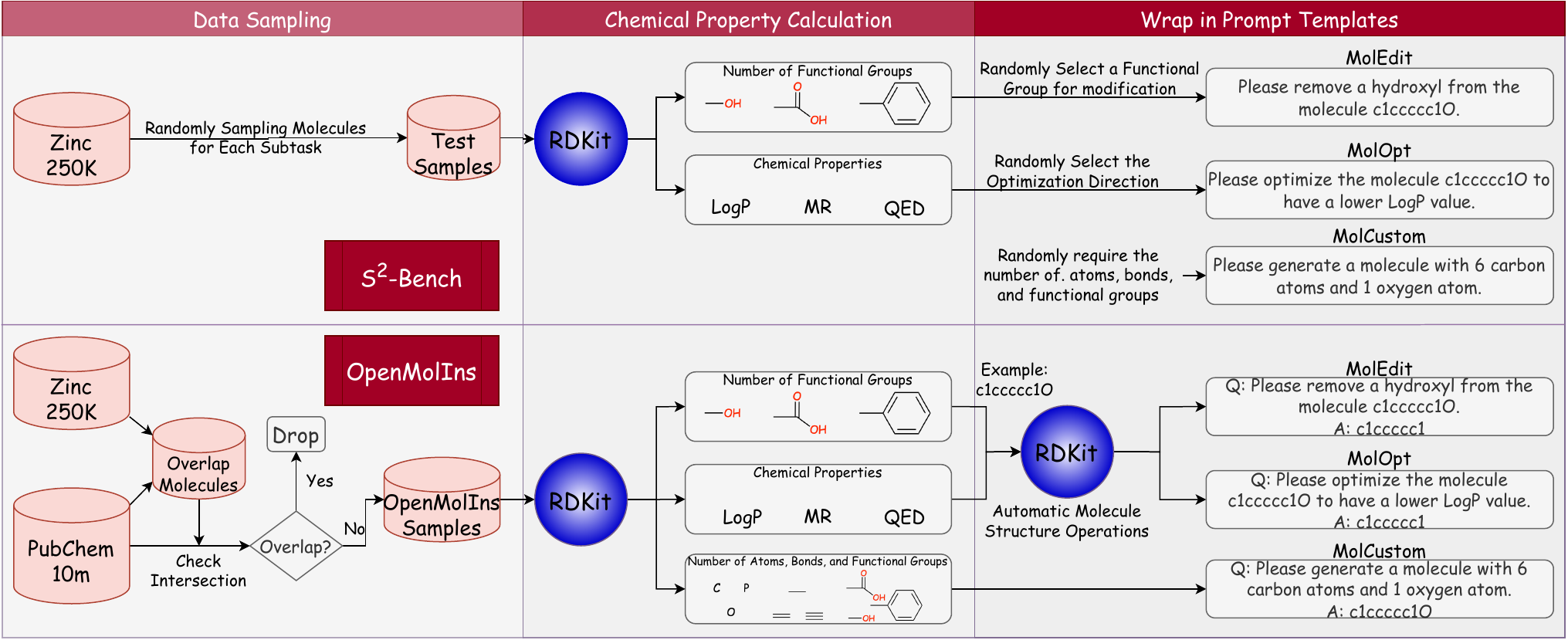}
    \caption{Data construction workflow of S\textsuperscript{2}-Bench \& OpenMolIns.}
    \Description{}
    \label{fig:data}
\end{figure*}

To effectively train and evaluate LLMs on the open-ended challenges posed by S\textsuperscript{2}-Bench, we introduce \textbf{OpenMolIns}, a specialized instruction-tuning dataset derived from the PubChem database. The core philosophy behind OpenMolIns is to provide models with the kind of flexible, one-to-many training examples they need to learn genuine chemical reasoning, rather than simply memorizing specific input-output pairs. To ensure the integrity of our evaluations, we meticulously designed OpenMolIns to have zero overlap with Zinc-250K, preventing any data leakage that could compromise the validity of our benchmark results. Detailed Comparison of the data construction process between S\textsuperscript{2}-Bench and OpenMolIns is shown in Figure \ref{fig:data}.

We built the instruction-tuning dataset programmatically, creating equal numbers of samples (i.e., 5,000) for all nine subtasks. The RDKit toolbox was essential to this process, allowing us to automatically generate data based on objective molecular properties and rules. This contrasts sharply with previous methods that rely on scarce and costly human annotations.

For \textbf{MolEdit} and \textbf{MolOpt}, our goal is to improve the capabilities of LLMs to perform rational, goal-oriented modifications. For MolEdit, we start with a molecule, programmatically identify a modifiable functional group, and then perform a change (add, delete, or substitute) to create a new, valid molecule. This pair of (original and modified) molecules is then used to construct an instruction. This process teaches the model not only a single correct answer, but rather the skill of modifying a molecule in a chemically sound way. Similarly, for MolOpt, we calculate properties before and after a modification to identify a specific optimization direction (e.g., increasing LogP or decreasing MR). The training sample then explicitly links a desired outcome to a structural change, teaching the LLM to reason about the relationship between structure and property.

In the \textbf{MolCustom} domain, we aim to train the model to generate molecules from scratch based on structural details. Instead of relying on existing molecule-caption pairs, we programmatically analyze molecules from PubChem, extract key structural statistics (e.g., atom counts, bond numbers, and functional group types), and use these as the basis for a training prompt. The molecule itself becomes the target output. This approach trains the model to synthesize a new molecule that fits a set of high-level, multi-faceted criteria, a skill fundamentally different from recalling a specific, pre-existing structure. This method allows the generated molecules to better fit the real distribution of molecular space and prepares the model for de novo design.

To further investigate the impact of data scales on an LLM's ability to learn these tasks, we created five distinct data levels: light, small, medium, large, and xlarge, shown in Table \ref{tab:sta}. This tiered structure allows researchers to systematically analyze the data scaling law for open-domain natural language-driven molecule generation, providing valuable insights into how data quantity influences a model's capacity for chemical reasoning.

\begin{table}[htbp]
    \centering
    \caption{Statistics of S\textsuperscript{2}-Bench and OpenMolIns.}
    \resizebox{1.0\columnwidth}{!}{
    \begin{tabular}{l|cc|ccccc}
    \toprule
    \rowcolor{gray!20} \textbf{Dataset} &  \multicolumn{2}{c}{\textbf{S\textsuperscript{2}-Bench}} & \multicolumn{5}{|c}{\textbf{OpenMolIns}} \\
    \midrule
     Item & each subtask & total & light & small & medium & large & xlarge \\
    \hline
    Data Size & 5,000 & 45, 000 & 4,500 & 18,000 & 45,000 & 90,000 & 1,200,000 \\
    \bottomrule
    \end{tabular}
    }
    
    \label{tab:sta}
\end{table}

\subsection{Statistics}

Here, we provide a detailed overview of S\textsuperscript{2}-Bench and OpenMolIns, as summarized in Tables \ref{tab:sta} and \ref{tab:comp}. The design and scale of our datasets are not arbitrary; they are meticulously crafted to enable a rigorous and nuanced evaluation of LLMs in open-domain molecule generation, which previous, smaller datasets cannot provide.

S\textsuperscript{2}-Bench is composed of nine subtasks, with each containing 5,000 test samples, for a total of 45,000 carefully curated test cases. This extensive size is critical for providing a statistically robust and comprehensive assessment of model performance, mitigating the risk of misleading results that can arise from smaller test sets. More importantly, these samples are designed to test the model's ability to handle one-to-many relationships, a challenge that is central to true chemical reasoning.

Meanwhile, to support the development of models capable of succeeding on this benchmark, OpenMolIns is provided in five distinct data scales, ranging from 4,500 to 1,200,000 examples. This tiered structure serves a dual purpose: it not only offers ample data to train powerful models but also allows researchers to systematically investigate the data scaling law for this novel task. By observing how model performance improves with increasing data size, we can gain valuable insights into the data requirements for aligning molecular space with natural language.

\begin{table}[htbp]
    \centering
    \caption{Comparison with existing text-based molecule generation benchmarks and datasets. Notably, we denote the statistics of the description-guided molecule design part in Mol-Instructions \cite{fang2023mol}.}
    \resizebox{1.0\columnwidth}{!}{
    \begin{tabular}{lllll}
    \toprule
    \rowcolor{gray!20} \textbf{Benchmarks or Datasets} & \textbf{Task Type} & \textbf{\# Training Data} & \textbf{\# Test Data} & \textbf{Public Access}\\
    \midrule
    PCdes\citep{zeng2022deep} & Targeted Generation & 10,500 & 3000 & $\times$\\
    ChEBI-20 \citep{edwards2022translation}& Targeted Generation & 26,407 & 3,000 & \checkmark\\
    PubChem324K \citep{liu2023molca} & Targeted Generation & 12,000& 2,000& \checkmark\\
    Mol-Instructions \citep{fang2023mol} & Targeted Generation & 297,319 & 1000 & \checkmark\\
    L + M - 24 \citep{edwards2024l+} & Targeted Generation & 160,492 & 21,839 & \checkmark\\
    \midrule
    \textbf{S\textsuperscript{2}-Bench \& OpenMolIns} & Open Generation & 1,200,000 & 45,000 & \checkmark\\
    \bottomrule
    \end{tabular}
    }

    \label{tab:comp}
\end{table}

Besides, as shown in Table \ref{tab:comp}, S\textsuperscript{2}-Bench and OpenMolIns stand out from existing text-based molecule generation datasets. S\textsuperscript{2}-Bench is the first to introduce a truly open-domain natural language-driven molecule generation task, moving the field beyond the limitations of one-to-one mapping.
In terms of data scale, our datasets are also significantly larger. OpenMolIns has the largest training set to date, with a volume that is unprecedented in this domain. This scale is crucial for enabling LLMs to learn complex chemical rules and generalize beyond the memorization of specific examples. Similarly, S\textsuperscript{2}-Bench's 45,000 test samples make it the largest evaluation benchmark, providing a more reliable and comprehensive measure of a model's true capabilities. This combination of innovative task design and large-scale data represents a new paradigm for evaluating and advancing LLMs in the field of text-guided molecular discovery.

We also report molecular distribution statistics for Zinc-250K, PubChem, and our pre-selected samples used in MolEdit and MolOpt. Specifically, we compare the average (i) atom count, (ii) ring count, (iii) branch count, and (iv) path length across datasets.

\begin{table}[htbp]
    \centering
    \caption{Molecular Distribution Statistics.}
    \resizebox{1.0\columnwidth}{!}{
    \begin{tabular}{l|c|c|c|c|c}
    \toprule
    \rowcolor{gray!20}\textbf{Dataset} & \textbf{\# Samples} & \textbf{Avg Atom Count} & \textbf{Avg Ring Count} & \textbf{Avg Branch Count} & \textbf{Avg Path Length} \\ 
    \midrule
    Zinc-250K & 250,000 & 23.15 & 2.76 & 7.31 & 12.48 \\
    PubChem & 9,000,000 & 25.18 & 2.8 & 7.87 & 13.07 \\
    S²-Bench's sampling & 30,000 & 23.17 & 2.74 & 7.34 & 12.47 \\ \bottomrule
    \end{tabular}
    }
    
    \label{tab:data_stat}
\end{table}

As shown in Table \ref{tab:data_stat}, Zinc-250K exhibits broadly similar statistical trends to PubChem, suggesting that Zinc-250K captures the core characteristics of drug-like chemistry and remains a reasonable proxy for discovery-scale small molecules. Meanwhile, the distribution of our pre-selected samples aligns closely with the original Zinc-250K, suggesting that our pre-selection does not substantially distort the underlying molecular distribution.
\section{Experiments}
\label{sec:Experiments}
\subsection{Models}
The models benchmarked are categorized into four groups: proprietary models, open-source general LLMs, open-source ChEBI-20 fine-tuned LLMs, and OpenMolIns fine-tuned LLMs.

\noindent\textbf{Proprietary Models.} This category includes LLMs that are only accessible via commercial API services. In this work, we benchmark GPT-4o, GPT-4-turbo, GPT-3.5-turbo \citep{achiam2023gpt}, Claude-3.5 \citep{claude3.5}, Claude-3 \citep{claude3}, and Gemini-1.5-pro \citep{gemini}, which are all the most advanced LLMs with powerful reasoning and generalization capabilities. These LLMs are pre-trained on vast pre-training corpora, normally including chemical-related knowledge.

\noindent\textbf{Open-source General \& Chemical LLMs.} This group contains open-source general LLMs with instruction following capability for a wide range of tasks, including Llama3-70B-Instruct, Llama3-8B-Instruct, Llama3.1-8B-Instruct, Llama3.2-1B-Instruct \citep{dubey2024llama}, Mistral-7B-Instruct-v0.2 \citep{jiang2023mistral}, Deepseek-R1-distill-Qwen-7B~\citep{guo2025deepseek}, Qwen2-7B-Instruct \citep{yang2024qwen2}, yi-1.5-9B \citep{young2024yi}, chatglm-9B \citep{glm2024chatglm}, Gemma3-12B \citep{team2025gemma}. Additionally, domain-specific LLMs like ChemLLM-20B \citep{zhang2024chemllm}, SmileyLlama \citep{cavanagh2026smileyllama},and ChemDFM-v1.5-8B \citep{zhao2025developing} are also considered.

\noindent\textbf{Open-source ChEBI-20 Fine-tuned LLMs.} LLMs fine-tuned on the ChEBI-20 dataset can grasp some extent of text-based molecule generation capability. In this case, our experiments also cover LLMs like MolT5-small, MolT5-base, MolT5-large \citep{edwards2022translation}, and BioT5-base \citep{pei2023biot5}. These models have been the state-of-art models in the molecule-caption translation task. Here, to ensure a fair evaluation of the intrinsic capabilities of LLMs, we exclude models that utilize multi-modal architectures or retrieval-augmented generation.

\noindent\textbf{OpenMolIns Fine-tuned LLMs.} We further fine-tune LLMs like Galactica-125M \citep{taylor2022galactica}, Llama3.2-1B-Instruct, and Llama3.1-8B-Instruct on OpenMolIns dataset for comparison. We specifically include the experiments on five distinct data sizes of OpenMolIns for Galactica-125M because the model has been pre-trained on scientific corpora and has proved its effectiveness in molecule-related tasks \citep{liu2023molca}. Meanwhile, a small-sized model can also help us study the data scaling law within a reasonable budget. Furthermore, Llama3.2-1B-Instruct and Llama3.1-8B-Instruct are also selected due to their advanced capabilities and similar architectures.

\subsection{Hyperparameters}

\label{sec:hyper}
We illustrate the detailed parameters adopted in our work, which is shown in Table \ref{tab:hyper}. Notably, we utilize one NVIDIA A100 80G for testing open-source LLMs, and 4$\times$NVIDIA A100 80G for instruction tuning. For closed-source LLMs, we call their official APIs.

\begin{table}[htbp]
    \centering
    \caption{Hyper-parameters.}
    \resizebox{0.4\columnwidth}{!}{
    \begin{tabular}{lc}
    \toprule
    \textbf{Item} & \textbf{Value} \\
    \midrule
    \rowcolor{gray!20} \textit{Generation} & \\
    temperature & 0.75 \\
    top\_p & 0.85 \\
    num\_beams & 1 \\
    max\_new\_tokens & 512 \\
    \hline
    \rowcolor{gray!20} \textit{Instruction Tuning} & \\
    epochs(light) & 10 \\
    epochs(small, medium) & 5 \\
    epochs(large, xlarge) & 3 \\
    batchsize & 32 \\
    lr & 3e-4 \\
    warmup\_ratio & 0.03 \\
    cutoff\_len & 1024 \\
    \hline
    \rowcolor{gray!20} \textit{Lora Settings} & \\
    r & 64 \\
    $\alpha$ & 128 \\
    dropout & 0.1 \\
    \bottomrule
    \end{tabular}
    }
    
    \label{tab:hyper}
\end{table}

\begin{figure*}
    \centering
    \includegraphics[width=0.9\linewidth]{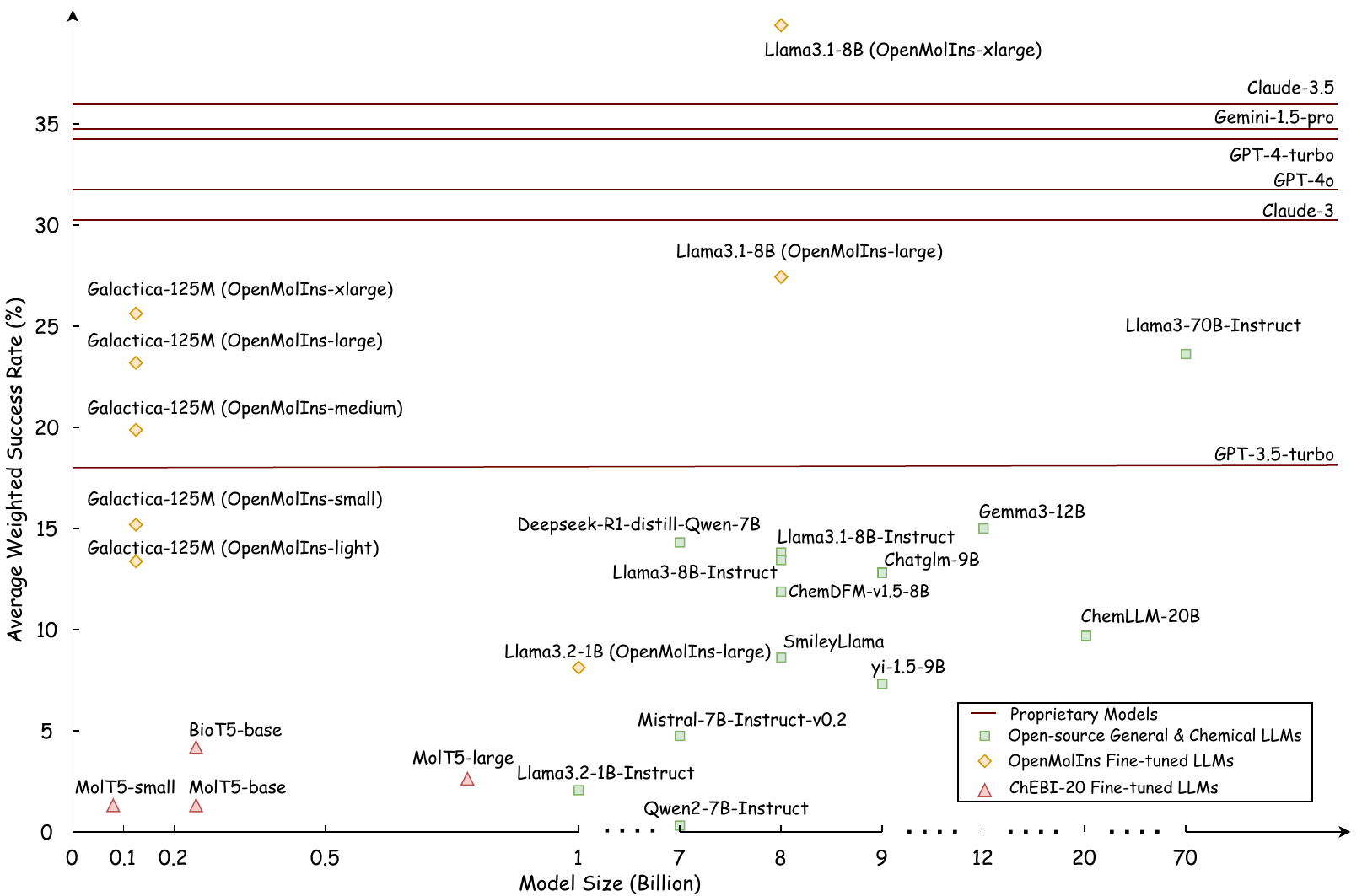}
    \caption{\textbf{The performance of LLMs evaluated in S\textsuperscript{2}-Bench.} LLMs fall into 4 categories: Proprietary Models, Open-source General LLMs, Open-source ChEBI-20 Fine-tuned LLMs, and OpenMolIns Fine-tuned LLMs. Models with unknown parameters are denoted as horizontal lines. }
    \Description{}
    \label{fig:per}
\end{figure*}

\subsection{Findings}
\label{sec:findings}
Based on our comprehensive benchmarking, we have made the following key observations regarding the capabilities of LLMs in open-domain natural language-driven molecule generation.

\noindent\textbf{F1: Current LLMs show promise but lack the structural understanding necessary for precise molecule generation.}
As shown in Figure \ref{fig:per}, top-tier proprietary models such as Claude-3.5 and Gemini-1.5-Pro attain 35.92\% and 34.80\% in average weighted success rate, respectively.
These results suggest that general LLMs hold promise for open-ended molecule generation due to their powerful reasoning capability, yet substantial challenges remain in translating textual requirements into chemically valid and structurally precise molecules.
Notably, within the MolCustom task, no LLM manages to surpass a 25\% weighted success rate on any individual subtask.
The particularly poor performance on MolCustom tasks highlights that while current LLMs can generate chemically plausible molecules in a broad sense, they struggle with satisfying fine-grained structural constraints such as precise atom, bond, or functional group counts, because the precise control of molecular structures requires reasoning over discrete, globally coupled constraints. LLMs, even trained on SMILES, primarily learn local token distributions from large-scale pre-training data, which rarely emphasizes exact numeric constraints.

The precise control, in fact, is crucial for molecular discovery, as many properties of interest, such as solubility, toxicity, and active site composition, depend on specific structural configurations.
Our findings suggest that open-domain natural language–driven molecule generation task can serve as an effective diagnostic tool, revealing systematic weaknesses in current LLMs and providing rich failure cases that can inform future instruction tuning or reinforcement learning strategies to improve molecule structural constraint adherence.


\noindent\textbf{F2: While pre-training provides a foundation for LLM performance, instruction tuning is indispensable for guiding and optimizing their capabilities in molecular discovery.} 
Open-source general LLMs can often lack chemical corpora during their pre-training stage. As a result, models like Qwen2-7B-Instruct and yi-1.5-9B obtained unsatisfactory results.
The observed shortcoming can be attributed to the lack of chemical data in pre-training, despite their notable proficiency in mathematics and general language understanding. Surprisingly, open-source chemical LLMs such as ChemLLM-20B and ChemDFM-v1.5-8B achieve relatively poor performance, indicating that current chemical LLM pre-training still fails to capture the molecular structural details.
In contrast, Llama3.1-8B-Instruct delivers superior performance at a relatively modest model scale, demonstrating a more balanced and comprehensive capability across diverse domains.

Meanwhile, our experiments with OpenMolIns highlight the effectiveness of instruction tuning. Fine-tuned on the largest OpenMolIns scale (i.e., xlarge), Llama3.1-8B-Instruct surpassed all other LLMs with an average weighted success rate of 39.33\%. More remarkably, the much smaller model, Galactica-125M, within just 125 million parameters, achieved an average weighted success rate of 25.73\%, outperforming models two orders of magnitude larger, such as Llama3-70B-Instruct. This remarkable outcome indicates that task-specific instruction tuning on a large, high-quality corpus can be highly effective, suggesting a practical avenue for LLMs to efficiently narrow the performance gap in molecular discovery.

\noindent\textbf{F3: One-to-one mappings between language instructions and molecules make LLMs inherently rely on pattern recognition and recall.}
Our results also substantiate the observation, demonstrating that existing one-to-one mappings are insufficient for evaluating true chemical comprehension of LLMs. 
Notably, ChEBI-20 Fine-tuned LLMs, which are designed to align molecules with texts, perform worse than general LLMs.
For instance, BioT5-base \citep{pei2023biot5}, a state-of-the-art model on the ChEBI-20 dataset, achieves an average weighted success rate of only 4.21\% on S\textsuperscript{2}-Bench.

A closer examination of BioT5-base’s performance provides insight into this limitation. 
Although the model achieves relatively high success rates on MolEdit and MolOpt tasks, its generated molecules often exhibit low similarity to the original structures. This indicates that the model is not performing rational edits; rather, it generates different but valid molecules that happen to satisfy the criteria.
In other words, BioT5-base relies more on recalling patterns from its training corpus than on learning to systematically modify molecular structures. Consequently, while the ChEBI-20 dataset is valuable for molecule–caption translation, its limited diversity, scale, and one-to-one mapping nature are insufficient for training LLMs to perform nuanced, open-ended molecular generation.

\begin{table*}[htbp]
    \centering
    \caption{Comparing WSR and SR with Human Expert Evaluation across BioT5-base and
             Galactica-125M. WSR consistently achieves stronger alignment with expert
             judgment than SR alone. Spearman correlations are computed over the six
             subtasks per model; $^{*}p < 0.05$.}
    \resizebox{0.7\textwidth}{!}{
    \begin{tabular}{l|l|cc|cc|cc}
    \toprule
        \rowcolor{gray!20}
        & &
        \multicolumn{2}{c|}{\textbf{BioT5-base}} &
        \multicolumn{2}{c|}{\textbf{Galactica-125M}} &
        \multicolumn{2}{c}{\textbf{Human Expert Eval.}} \\
        \rowcolor{gray!20}
        \textbf{Task} & \textbf{Subtask} &
        \textbf{SR} & \textbf{WSR} &
        \textbf{SR} & \textbf{WSR} &
        \textbf{BioT5-base} & \textbf{Galactica-125M} \\
        \midrule
        ~ & AddComponent & 0.3462 & 0.0542 & 0.5842 & 0.3423 & 0.64 & 3.57 \\
        MolEdit & DelComponent & 0.1668 & 0.0266 & 0.6526 & 0.3318 & 0.22 & 3.00 \\
        ~ & SubComponent & 0.0684 & 0.0108 & 0.1872 & 0.1128 & 0.19 & 2.60 \\
        \midrule
        ~ & LogP & 0.5158 & 0.0787 & 0.7362 & 0.4429 & 0.58 & 3.74 \\
        MolOpt & MR & 0.5060 & 0.0808 & 0.7124 & 0.4059 & 0.67 & 3.47 \\
        ~ & QED & 0.5068 & 0.0801 & 0.5786 & 0.3285 & 0.64 & 3.40 \\
        \midrule
        \rowcolor{gray!10}
        \multicolumn{2}{l|}{\textbf{Spearman $\rho$ (vs.\ Human Expert Eval.)}} &
        0.551 & $\mathbf{0.899}^{*}$ &
        0.714 & $\mathbf{0.886}^{*}$ &
        \multicolumn{2}{c}{---} \\
        \rowcolor{gray!10}
        \multicolumn{2}{l|}{\textit{p}-value} &
        0.257 & \textbf{0.015} &
        0.111 & \textbf{0.019} &
        \multicolumn{2}{c}{} \\
    \bottomrule
    \end{tabular}
    }
    \label{tab:human_combined}
\end{table*}

\begin{figure}[htbp]
    \centering
    \includegraphics[width=1.0\linewidth]{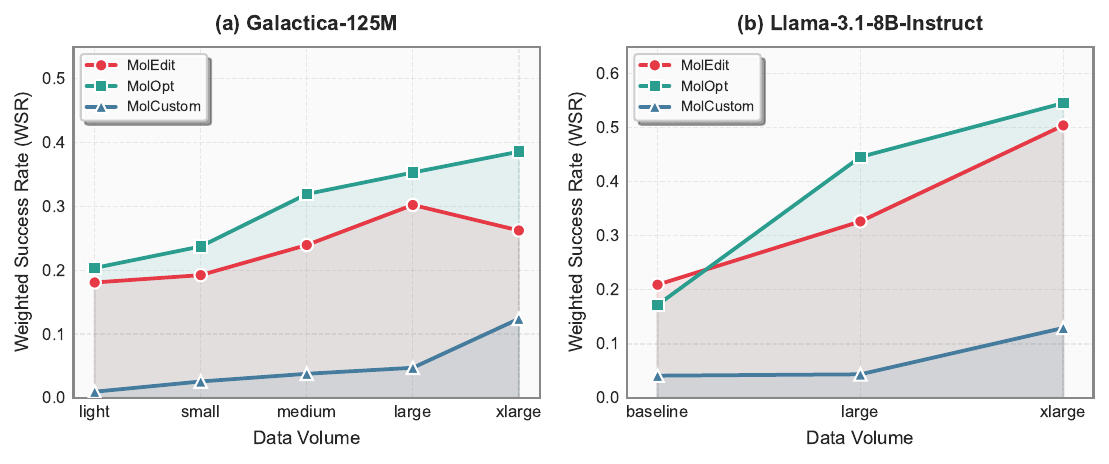}
    \caption{Task-specific performance scaling with increasing data in S\textsuperscript{2}-Bench. (a) Galactica-125M. (b) Llama3.1-8B-Instruct.}
    \Description{}
    \label{fig:scale}
\end{figure}
\noindent\textbf{F4: The data scaling law does not always hold: task-specific bottlenecks limit gains in molecule optimization and editing.}
As shown in Figure \ref{fig:scale}, our experiments with Galactica-125M across the five data scales provide new insights into the data scaling law for molecule generation. Overall, we observe that larger datasets generally improve performance, but the effect is highly task-dependent.

In \textbf{MolCustom}, scaling the dataset from large to xlarge yields a remarkable 286\% average performance gain, while there is limited performance improvement from light to large, indicating that tasks involving complex, de novo synthesis strongly benefits from larger and more diverse training data.
Conversely, Galactica-125M obtains only modest gains in \textbf{MolOpt}, and even negative gains in \textbf{MolEdit}, suggesting that additional data alone cannot overcome existing bottlenecks in these relatively simpler tasks.

Interestingly, when examining Llama3.1-8B-Instruct, we find that performance continues to increase when scaling from the large to xlarge data level, except for LogP and MR subtasks. This contrast generally highlights that small models like Galactica-125M are capacity-limited, whereas larger models can more effectively leverage additional data.

To summarize, these findings suggest that while complex, open-ended tasks are data-driven, simpler property optimization or editing tasks may be constrained more by model capacity than by dataset size. Thus, merely enlarging training data may not yield proportional benefits. Instead, future progress will require a balanced consideration of both model scale and data scale.

\subsection{Validating WSR with Human Evaluation}

We conducted an extra correlation analysis to justify our design for using similarity in Weighted Success Rate (WSR). Here, we focused on the unusual performance of BioT5-base in MolEdit and MolOpt, where the model achieves high success rates but performs poorly in the quality term, indicating that the generated molecules are not modified from the original molecule but from scratch. 

For this analysis, we first randomly sampled 100 examples from each subtask and then asked human experts to evaluate the correctness and quality of the generated molecules. Each example is assigned an expert score from 0 to 5, where:
\begin{itemize}
    \item 5 denotes a perfect success,
    \item 4 means success but with redundant operations (a little different structure),
    \item 3 shows failure but modified from the original molecule (similar structure),
    \item 2 denotes failure and a little different structure,
    \item 1 indicates success but an unrelated/different structure,
    \item 0 denotes failure and an unrelated/different structure.
\end{itemize}

We calculate the average score of expert evaluation. 
Table~\ref{tab:human_combined} presents per-subtask SR, WSR, and expert scores for
BioT5-base and Galactica-125M, with Spearman correlations reported at the bottom.
WSR achieves significantly stronger alignment with expert judgment than SR alone across
both models: for BioT5-base, $\rho_{\mathrm{WSR}} = 0.899$ ($p = 0.015$) versus
$\rho_{\mathrm{SR}} = 0.551$ ($p = 0.257$); for Galactica-125M,
$\rho_{\mathrm{WSR}} = 0.886$ ($p = 0.019$) versus $\rho_{\mathrm{SR}} = 0.714$
($p = 0.111$). In both cases WSR reaches statistical significance while SR does not.

Therefore, it is essential to apply similarities as weights of Success Rate to reflect the genuine capabilities of LLMs more faithfully.

The results confirm that SR is a noisy proxy for quality: as a binary criterion, it rewards any molecule satisfying the target constraint irrespective of structural distance from the input, allowing models to score high SR by generating chemically valid but structurally unrelated outputs, a pattern human experts consistently penalize.
WSR corrects for this by down-weighting successes achieved through large structural departures, and the significantly higher Spearman correlations confirm that this
re-weighting tracks human judgment more faithfully.

\subsection{Extending MolOpt to Additional Properties}

The choice of LogP, MR, and QED in MolOpt was motivated by the need for automated, reproducible, and unambiguous evaluation at scale, which currently rules out many complex properties. To address the concern that this set may be overly simplistic, we extend MolOpt with two more challenging properties: \textbf{Topological Polar Surface Area (TPSA)}, widely used to estimate drug permeability, and \textbf{HOMO-LUMO Gap (HLG)}, a fundamental quantum-chemical descriptor of molecular reactivity. Due to resource constraints, we report results on a set of representative models in Table~\ref{tab:molopt_extended}. The overall success rates on TPSA and HLG are noticeably lower than those on LogP/MR/QED, confirming that these properties pose a harder optimization challenge and that MolOpt is far from saturated; meanwhile, no single model dominates across both properties, suggesting complementary strengths on different property axes and reinforcing the value of MolOpt as a discriminative benchmark that naturally extends beyond the original three properties.

\begin{table}[htbp]
\centering
\caption{Extended MolOpt results on TPSA and HOMO-LUMO Gap (HLG). SR denotes Success Rate and WSR denotes Weighted Success Rate.}
\label{tab:molopt_extended}
\resizebox{1.0\columnwidth}{!}{
\begin{tabular}{lcccc}
\toprule

\rowcolor{gray!20}\textbf{Model} & \textbf{TPSA SR} & \textbf{TPSA WSR} & \textbf{HLG SR} & \textbf{HLG WSR} \\
\midrule
MolT5-large            & 0.3568 & 0.0347 & 0.3460 & 0.0319 \\
ChemDFM-v1.5-8B        & 0.0634 & 0.0440 & 0.0920 & 0.0578 \\
Llama3.1-8B-Instruct   & 0.3654 & 0.1745 & 0.1760 & 0.0881 \\
\bottomrule
\end{tabular}
}
\end{table}

\section{Conclusion}
\label{sec:conclusion}
In this study, we introduced \textbf{S\textsuperscript{2}-Bench} and \textbf{OpenMolIns}, the \textbf{first} benchmark and instruction tuning dataset for evaluating the capabilities of LLMs in open-domain natural language-driven molecule generation. By moving beyond traditional one-to-one text-to-molecule mappings, our benchmark accommodates the open-ended nature of molecular discovery. S\textsuperscript{2}-Bench focuses on the realistic molecular reasoning and design, rather than simple pattern matching and memorization.
Our comprehensive benchmarking of 31 LLMs not only highlights the challenges faced by current models and the limitations of existing targeted molecule generation datasets, but also demonstrates the substantial potential of LLMs for natural language–driven molecule discovery.

\newpage
\section*{Acknowledgments}
This research was supported by the National Natural Science Foundation of China (Grant No.: T2541073, 62372314), the Hong Kong Research Grants Council under the Theme-based Research Scheme (project no. T41-517/25-N), and the NSFC/RGC Joint Research Scheme (N\_PolyU5179/25). The experimental part of this work was supported by The Centre for Large AI Models (CLAIM) of The Hong Kong Polytechnic University.

\bibliographystyle{ACM-Reference-Format}
\bibliography{sample-base}

\appendix

\section{Testing Process}
\label{sec:test}
In this section, we present the testing process used to calculate the success rate of each subtask.

\begin{algorithm}
\caption{MolEdit Testing Process}
\renewcommand{\algorithmicrequire}{\textbf{Input:}}
\renewcommand{\algorithmicensure}{\textbf{Output:}}
\label{alg:edit}
\begin{algorithmic}
\REQUIRE Generated Molecule $m^g$, Original Molecule $m^o$, Subtask $t$, Group-to-add $\alpha$, Group-to-delete $\Delta$
\ENSURE Pass or Not (\textbf{Bool})
    
\algorithmicif\ $t$ is AddComponent\ \algorithmicthen 

\ \ \ \ \algorithmicif\ $|\alpha|$($m^g$) = $|\alpha|$($m^o)$ + 1\ \algorithmicthen 

\ \ \ \ \ \ \ \ return \textbf{true} 

\ \ \ \ \algorithmicelse 

\ \ \ \ \ \ \ \ return \textbf{false} 

\algorithmicelse\ \algorithmicif\ $t$ is DelComponent\ \algorithmicthen 

\ \ \ \ \algorithmicif\ $|\Delta|$($m^g$) = $|\Delta|$($m^o)$ - 1\ \algorithmicthen 

\ \ \ \ \ \ \ \ return \textbf{true} 

\ \ \ \ \algorithmicelse 

\ \ \ \ \ \ \ \ return \textbf{false} 

\algorithmicelse\ \algorithmicif\ $t$ is SubComponent\ \algorithmicthen 

\ \ \ \ \algorithmicif\ $|\Delta|$($m^g$) = $|\Delta|$($m^o)$ - 1\ \algorithmicand\ $|\alpha|$($m^g$) = $|\alpha|$($m^o)$ + 1\ \algorithmicthen 

\ \ \ \ \ \ \ \ return \textbf{true} 

\ \ \ \ \algorithmicelse 

\ \ \ \ \ \ \ \ return \textbf{false}
\end{algorithmic}

\end{algorithm}

\begin{algorithm}
\caption{MolOpt Testing Process}
\renewcommand{\algorithmicrequire}{\textbf{Input:}}
\renewcommand{\algorithmicensure}{\textbf{Output:}}
\label{alg:opt}
\begin{algorithmic}
\REQUIRE Generated Molecule $m^g$, Original Molecule $m^o$, Subtask $t$, Optimization Direction Requirement $R$

\ENSURE Pass or Not (\textbf{Bool})

\algorithmicif\ $t$ is LogP\ \algorithmicthen 

\ \ \ \ \algorithmicif\ $LogP$($m^g$) $>$ $LogP$($m^o)$\ \algorithmicand\ R is $\uparrow$\ \algorithmicthen 

\ \ \ \ \ \ \ \ return \textbf{true} 

\ \ \ \ \algorithmicelse\ \algorithmicif\ $LogP$($m^g$) $<$ $LogP$($m^o)$\ \algorithmicand\ R is $\downarrow$\ \algorithmicthen 

\ \ \ \ \ \ \ \ return \textbf{true} 

\ \ \ \ \algorithmicelse 

\ \ \ \ \ \ \ \ return \textbf{false} 

\algorithmicelse\ \algorithmicif\ $t$ is MR\ \algorithmicthen 

\ \ \ \ \algorithmicif\ $M\!R$($m^g$) $>$ $M\!R$($m^o)$\ \algorithmicand\ R is $\uparrow$\ \algorithmicthen 

\ \ \ \ \ \ \ \ return \textbf{true} 

\ \ \ \ \algorithmicelse\ \algorithmicif\ $M\!R$($m^g$) $<$ $M\!R$($m^o)$\ \algorithmicand\ R is $\downarrow$\ \algorithmicthen 

\ \ \ \ \ \ \ \ return \textbf{true} 

\ \ \ \ \algorithmicelse 

\ \ \ \ \ \ \ \ return \textbf{false} 

\algorithmicelse\ \algorithmicif\ $t$ is QED\ \algorithmicthen 

\ \ \ \ \algorithmicif\ $Q\!E\!D$($m^g$) $>$ $Q\!E\!D$($m^o)$\ \algorithmicand\ R is $\uparrow$\ \algorithmicthen 

\ \ \ \ \ \ \ \ return \textbf{true} 

\ \ \ \ \algorithmicelse\ \algorithmicif\ $Q\!E\!D$($m^g$) $<$ $Q\!E\!D$($m^o)$\ \algorithmicand\ R is $\downarrow$\ \algorithmicthen 

\ \ \ \ \ \ \ \ return \textbf{true} 

\ \ \ \ \algorithmicelse 

\ \ \ \ \ \ \ \ return \textbf{false}
\end{algorithmic}
\end{algorithm}

\begin{algorithm}
\caption{MolCustom Testing Process}
\renewcommand{\algorithmicrequire}{\textbf{Input:}}
\renewcommand{\algorithmicensure}{\textbf{Output:}}
\label{alg:custom}
\begin{algorithmic}
\REQUIRE Generated Molecule $m^g$, Subtask $t$, Atom List $A$, Bond List $B$, Functional Group List $G$, Requirements $R$

\ENSURE Pass or Not (\textbf{Bool})

flag = \algorithmictrue

\algorithmicif\ $t$ is AtomNum\ \algorithmicthen

\ \ \ \ \algorithmicfor\ $atom$\ in\ $A$\ \textbf{do}

\ \ \ \ \ \ \ \ \algorithmicif\ $|atom|$($m^g$) $\neq$ $R$[$atom$]\ \algorithmicthen 

\ \ \ \ \ \ \ \ \ \ \ \ flag = \textbf{false}  

\algorithmicelse\ \algorithmicif\ $t$ is BondNum\ \algorithmicthen 

\ \ \ \ \algorithmicfor\ $bond$\ in\ $B$\ \textbf{do}

\ \ \ \ \ \ \ \ \algorithmicif\ $|bond|$($m^g$) $\neq$ $R$[$bond$]\ \algorithmicthen 

\ \ \ \ \ \ \ \ \ \ \ \ flag = \textbf{false}  

\algorithmicelse\ \algorithmicif\ $t$ is FunctionalGroup\ \algorithmicthen 

\ \ \ \ \algorithmicfor\ $group$\ in\ $G$\ \textbf{do}

\ \ \ \ \ \ \ \ \algorithmicif\ $|group|$($m^g$) $\neq$ $R$[$group$]\ \algorithmicthen 

\ \ \ \ \ \ \ \ \ \ \ \ flag = \textbf{false}  

return flag
\end{algorithmic}
\end{algorithm}

\section{Detailed Results}
\label{sec:res}

In this section, we show the leaderboard of S\textsuperscript{2}-Bench in Table \ref{tab:rank}. The detailed experimental results of all the subtasks in Table \ref{tab:moledit}, \ref{tab:molopt}, and \ref{tab:molcustom}. Note that \textbf{we do not provide results of vanilla galactica-125M, for it's not instruction-tuned and cannot respond to human instructions.} All results of vanilla galactica-125M should be close to zero.

\begin{table}[htbp]
    \centering
    \caption{Leaderboard of S\textsuperscript{2}-Bench.}
    \resizebox{0.85\columnwidth}{!}{
    \begin{tabular}{ccccc}
    \toprule
    \rowcolor{gray!20}\textbf{Model} & \textbf{\# Parameters (B)} & \textbf{$\overline{S\!R}$ (\%)} & \textbf{$\overline{W\!S\!R} (\%)$} & \textbf{Rank} \\
    \midrule
    Llama3.1-8B (OpenMolIns-xlarge) & 8 & 58.79 &39.33 & 1\\
    Claude-3.5 \citep{claude3.5} & - & 51.10 & 35.92 & 2\\
    Gemini-1.5-pro \citep{gemini} & - & 52.25 & 34.80 & 3\\
    GPT-4-turbo \citep{achiam2023gpt} & - & 50.74 & 34.23 & 4\\
    GPT-4o \citep{achiam2023gpt} & - & 49.08 & 32.29 & 5\\
    Claude-3 \citep{claude3} & - & 46.14 & 30.47 & 6 \\
    Llama3.1-8B (OpenMolIns-large)& 8 & 43.1 & 27.22 & 7 \\
    Galactica-125M (OpenMolIns-xlarge)& 0.125 & 44.48 & 25.73 & 8 \\
    Llama3-70B-Instruct (Int4) \citep{dubey2024llama}& 70 & 38.54 & 23.93 & 9\\
    Galactica-125M (OpenMolIns-large) & 0.125 & 39.28 & 23.42 & 10 \\
    Galactica-125M (OpenMolIns-medium) & 0.125 & 34.54 & 19.89 & 11 \\
    GPT-3.5-turbo \citep{achiam2023gpt} & - & 28.93 & 18.58 & 12\\
    Galactica-125M (OpenMolIns-small) & 0.125 & 24.17 & 15.18 & 13\\
    Gemma3-12B \citep{team2025gemma} & 12 & 26.28& 15.00 & 14\\
    Deepseek-R1-distill-Qwen-7B~\citep{guo2025deepseek} & 7 & 25.07 & 14.61 & 15\\
    Llama3.1-8B-Instruct \citep{dubey2024llama} & 8 & 26.12 & 13.78 & 16\\
    Llama3-8B-Instruct \citep{dubey2024llama} & 8 & 26.40 & 13.75 & 17\\
    chatglm-9B \citep{glm2024chatglm} & 9 & 18.50 & 13.13(7) & 18 \\
    Galactica-125M (OpenMolIns-light) & 0.125 & 20.95 & 13.13(6) &19\\
    ChemDFM-v1.5-8B \citep{zhao2025developing}&	8 & 18.24	&12.07	&20 \\
ChemLLM-20B \citep{zhang2024chemllm} & 20 &	16.23	&9.76	&21\\
    SmileyLlama \citep{cavanagh2026smileyllama} & 8 & 16.18 & 8.43 & 22 \\
    Llama3.2-1B (OpenMolIns-large) & 1 & 14.11 & 8.10 & 23\\
    yi-1.5-9B \citep{young2024yi}& 9 & 14.10 & 7.32 & 24\\
    Mistral-7B-Instruct-v0.2 \citep{jiang2023mistral}& 7 & 11.17 & 4.81 & 25\\
    BioT5-base \citep{pei2023biot5}& 0.25 & 24.19& 4.21 &26\\
    MolT5-large \citep{edwards2022translation}& 0.78 & 23.11 & 2.89 & 27\\
    Llama3.2-1B-Instruct \citep{dubey2024llama}& 1 & 3.95 & 1.99 & 28\\
    MolT5-base \citep{edwards2022translation} & 0.25 & 11.11 & 1.30(0) & 29\\
    MolT5-small \citep{edwards2022translation} & 0.08 & 11.55 & 1.29(9) & 30\\
    Qwen2-7B-Instruct \citep{yang2024qwen2} & 7 & 0.18 & 0.15 & 31\\
    \bottomrule
    \end{tabular}
    }
    
    \label{tab:rank}
\end{table}

 \begin{table}[t]
     \centering
     \caption{Detailed results on MolEdit. For each task, we highlight the \colorbox{backred!50}{best} and the \colorbox{backblue!75}{second-best} success rate (SR), as well as the weighted success rate (WSR).}
     \resizebox{1.0\columnwidth}{!}{
     \begin{tabular}{c|cccc|cccc|cccc}
     \toprule
         \multirow{2}{*}{Models} & \multicolumn{4}{c|}{AddComponent} & \multicolumn{4}{c|}{DelComponent} & \multicolumn{4}{c}{SubComponent}\\ \cline{2-13}
          & SR & Similarity &WSR & Validity & SR & Similarity & WSR &Validity & SR & Similarity & WSR & Validity \\
     \midrule
          GPT-4o \citep{achiam2023gpt} & 0.6188&	0.6782& 0.4197&	0.7412&	0.7012&	0.6038& 0.4234 &	0.8474&	\colorbox{backblue!75}{0.7992} &	0.7225& \colorbox{backblue!75}{0.5774}&	0.9368 \\
          GPT-4-turbo \citep{achiam2023gpt} & 0.6990 &	0.6936& \colorbox{backblue!75}{0.4848}&	0.7934& 0.7244 &	0.5735& 0.4154 &0.9060&	0.7778&	0.7323& 0.5696&	0.9160 \\
          GPT-3.5-turbo \citep{achiam2023gpt} & 0.5832&	0.6545& 0.3817&	0.7980&	0.3082& 0.7797& 0.2403&	0.8468&	0.2918&	0.6333& 0.1848&	0.6822 \\
          Claude-3.5 \citep{claude3.5} & 0.6832&	0.7017& 0.4794 &	0.4414&	0.5414&	0.6678&	0.3615& 0.7960&	\colorbox{backred!50}{0.8104} &	0.7310&\colorbox{backred!50}{0.5924}&	0.9588 \\
          Claude-3 \citep{claude3} & 0.6766&	0.6840& 0.4628&	0.8180&	0.5556&	0.6408& 0.3560&	0.8984&	0.6550&	0.7159& 0.4689&	0.9184 \\
          Gemini-1.5-pro \citep{gemini} & \colorbox{backblue!75}{0.7058} &	0.6792& 0.4794 &	0.8254&	\colorbox{backblue!75}{0.7590} &	0.5949& \colorbox{backblue!75}{0.4515}&	0.9158&	0.7148&	0.7139& 0.5103&	0.8684 \\
          \midrule
          Llama3-70B-Instruct (Int4) \citep{dubey2024llama} & 0.5198&	0.6801& 0.3535&	0.5922&	0.6122&	0.5637& 0.3451&	0.7182&	0.5094&	0.7170& 0.3652&	0.6822 \\
          Llama3-8B-Instruct \citep{dubey2024llama} & 0.3914&	0.6649& 0.2602&	0.5374	&0.4348&	0.5058& 0.2199&	0.5700&	0.2602&	0.6841& 0.1780&	0.4838 \\
          Llama3.1-8B-Instruct \citep{dubey2024llama} & 0.2992&	0.6088& 0.1822&	0.4954 & 0.4416 & 0.5199 & 0.2296& 0.5832 &0.3401 &0.6424 & 0.2185& 0.5076 \\ 
          Mistral-7B-Instruct-v0.2 \citep{jiang2023mistral}& 0.1868&	0.6251&0.1168&	0.3760&	0.2018&	0.3774&	0.0762&0.3590&	0.0602&	0.6227&0.0375&	0.3550 \\
          Qwen2-7B-Instruct \citep{yang2024qwen2}& 0.0010&	0.2527& 0.0003&	0.0036&	0.0006&	0.4024& 0.0002&	0.0012&	0.0004&	0.2895& 0.0001&	0.0068 \\
          Yi-1.5-9B \citep{young2024yi}& 0.1742&	0.4170&0.0726&	0.4216&	0.2858&	0.5936&	0.1697&0.4909&	0.1370&	0.4619&0.0633&	0.4368 \\
          Chatglm-9B \citep{glm2024chatglm} & 0.2932&	0.7622&	0.2235& 0.5686&	0.2956&	0.7494&	0.2215&0.6914&	0.1498&	0.7150& 0.1071&	0.5084 \\
          Llama3.2-1B-Instruct \citep{dubey2024llama} & 0.0374&	0.5343& 0.0200&	0.1982&	0.0768&	0.5750&0.0442&	0.3028&	0.0102&	0.3671&0.0037&	0.1468\\
          Gemma3-12B \citep{team2025gemma}& 0.4300 & 0.7081& 0.3045 & 0.6324& 0.4948&	0.5158&	0.2552& 0.7168&	 	0.4208&	0.7154 & 0.3010 &	0.7616	 \\
          Deepseek-R1-distill-Qwen-7B~\citep{guo2025deepseek} & 0.2667 & 0.6904 & 0.1841 & 0.0090 & 0.6367 & 0.5921 & 0.3770 & 0.0736 & 0.2156 & 0.7380 & 0.1591 & 0.0356 \\
          SmileyLlama \citep{cavanagh2026smileyllama} & 0.1990 & 0.5460 & 0.1088 & 0.7700 & 0.0740 & 0.5990 & 0.0445 & 0.7940 & 0.0170 & 0.5870 & 0.0102 & 0.7740 \\
          ChemDFM-v1.5-8B \citep{zhao2025developing}&	0.2942&	0.6454&	0.1899&	0.9112&	0.2056&	0.6937&	0.1426&	0.9286&	0.2198&	0.6432&	0.1414&	0.8652 \\
          ChemLLM-20B \citep{zhang2024chemllm} & 0.1436&	0.5978&	0.0858&	0.2108&	0.2092&	0.5734&	0.1200&	0.3098&	0.3088&	0.6805&	0.2101&	0.4936\\
          \midrule
          MolT5-small \citep{edwards2022translation}& 0.1220&	0.1027& 0.0125&	0.4490&	0.1598&	0.1125&0.0180& 0.4504&	0.0708&	0.1029&	0.0073&0.4876 \\
          MolT5-base \citep{edwards2022translation}& 0.1354&	0.1066& 0.0144&	0.4686&	0.1562&	0.1144&	0.0179& 0.4472& 0.0584&	0.1028& 0.0060&	0.4426 \\
          MolT5-large \citep{edwards2022translation}& 0.2834&	0.1084& 0.0307&	0.9282&	0.2228&	0.1201& 0.0268&	0.9198&	0.1692&	0.0932& 0.0158&	0.9410 \\
          BioT5-base \citep{pei2023biot5} & 0.3462&	0.1567& 0.0542&	1.0000&	0.1668&	0.1597& 0.0266&	1.0000&	0.0684&	0.1576& 0.0108&	0.9998 \\
          \midrule
          Llama3.2-1B (OpenMolIns-large)& 0.1756&	0.5676&	0.0997& 0.3216&	0.1816&	0.4963&	0.0901& 0.2466&	0.0844&	0.5415& 0.0457&	0.2958\\
          Llama3.1-8B (OpenMolIns-large)& 0.5822&	0.6541& 0.3808&	0.6730&	0.5104&	0.5074&	0.2590& 0.6896&	0.5440&	0.6258& 0.3404&	0.8400 \\
          Llama3.1-8B (OpenMolIns-xlarge)& \colorbox{backred!50}{0.7790}&	0.6769& \colorbox{backred!50}{0.5273} &	0.9468&	\colorbox{backred!50}{0.8640}&	0.6166&	\colorbox{backred!50}{0.5327}& 0.9132&	0.6100&	0.7434& 0.4535&	0.9596 \\
          Galactica-125M (OpenMolIns-light) & 0.3786&	0.5958&	0.2256& 0.6842&	0.2062&	0.6521& 0.1345&	0.7048&	0.3102&	0.5879& 0.1824&	0.6674 \\
          Galactica-125M (OpenMolIns-small) & 0.3472&	0.6172& 0.2143&	0.5356&	0.3258&	0.6025& 0.1963&	0.5758&	0.2692&	0.6181& 0.1664&	0.5692 \\
          Galactica-125M (OpenMolIns-medium) & 0.4736&	0.5682&0.2691&	0.7442&	0.4886&	0.5184&	0.2533& 0.7488&	0.3282&	0.5975& 0.1961&	0.6958 \\
          Galactica-125M (OpenMolIns-large) & 0.5866&	0.5876& 0.3447&	0.8228&	0.6078&	0.5577& 0.3390&	0.7934&	0.3438&	0.6491& 0.2232&	0.8438 \\
          Galactica-125M (OpenMolIns-xlarge) & 0.5842&	0.5859& 0.3423&	0.8438&	0.6526&	0.5084&	0.3318& 0.8286&	0.1872&	0.6024& 0.1128&	0.8538 \\
     \bottomrule
     \end{tabular}
     }

     \label{tab:moledit}
 \end{table}

 \begin{table}[t]
     \centering
     \caption{Detailed results on MolOpt. For each task, we highlight the \colorbox{backred!50}{best} and the \colorbox{backblue!75}{second-best} success rate (SR), as well as the weighted success rate (WSR).}
     \resizebox{1.0\columnwidth}{!}{
     \begin{tabular}{c|cccc|cccc|cccc}
     \toprule
         \multirow{2}{*}{Models} & \multicolumn{4}{c|}{LogP} & \multicolumn{4}{c|}{MR} & \multicolumn{4}{c}{QED}\\ \cline{2-13}
          & SR & Similarity & WSR & Validity & SR & Similarity & WSR & Validity & SR & Similarity & WSR & Validity \\
     \midrule
          GPT-4o~\citep{achiam2023gpt} & 0.7190&	0.6586&0.4735&	0.8796&	0.6864&	0.6420 &0.4407	&0.8352&	0.3952&	0.6180& 0.2442&	0.8570 \\
          GPT-4-turbo~\citep{achiam2023gpt} & 0.7662&	0.6984& 0.5351&	0.9048&	\colorbox{backblue!75}{0.7388}&	0.6821& \colorbox{backblue!75}{0.5039}&	0.8848&	0.3946&	0.6587& 0.2599&	0.9050 \\
          GPT-3.5-turbo \citep{achiam2023gpt}& 0.4048&	0.6327& 0.2561&	0.8540&	0.4120&	0.6263&	0.2580& 0.8486&	0.3316&	0.5635&0.1869&	0.8354 \\
          Claude-3.5 \citep{claude3.5}& 0.7970&	0.7124& \colorbox{backblue!75}{0.5678}&	0.9422&	0.6962&	0.7112&0.4951&	0.9110&	0.5361&	0.7042&\colorbox{backblue!75}{0.3775}&	0.8604 \\
          Claude-3 \citep{claude3} & 0.7984 &	0.6067&	0.4844& 0.9096&	0.6094&	0.6398& 0.3899&	0.9062&	0.4678&	0.5855& 0.2739&	0.9044 \\
          Gemini-1.5-pro \citep{gemini}& 0.7712&	0.7022& 0.5415&	0.9274&	\colorbox{backred!50}{0.7876} &	0.6744&\colorbox{backred!50}{0.5312}&	0.8926&	0.4704&	0.6077&0.2859&	0.9484 \\
          \midrule
          Llama3-70B-Instruct (Int4) \citep{dubey2024llama} & 0.5984&	0.6028&0.3607&	0.6482&	0.5684&	0.6032&0.3429&	0.6272&	0.2774&	0.4828&0.1339&	0.6340 \\
          Llama3-8B-Instruct \citep{dubey2024llama}& 0.4642&	0.3658& 0.1698&	0.6086&	0.4332&	0.4793& 0.2076&	0.5704&	0.2568&	0.4547&0.1168&	0.6112 \\
          Llama3.1-8B-Instruct \citep{dubey2024llama}& 0.3990 & 0.4235& 0.1690&  0.5122& 0.4134 &  0.4794 &0.1982 & 0.5162 & 0.2655 & 0.4499 & 0.1194& 0.6158 \\
          Mistral-7B-Instruct-v0.2 \citep{jiang2023mistral}& 0.2220&	0.4501& 0.0999&	0.2802&	0.1908&	0.2578& 0.0492&	0.3795&	0.1210&	0.3244& 0.0393&	0.2532 \\
          Qwen2-7B-Instruct \citep{yang2024qwen2} & 0.0000&	0.2923& 0.0000&	0.0004&	0.0002&	0.4123& 0.0001&	0.0004&	0.0000	&0.0000&	0.0000&0.0000 \\
          Yi-1.5-9B \citep{young2024yi} & 0.2884&	0.5461& 0.1575 &	0.4927&	0.2050&	0.3724& 0.0763&	0.4126&	0.1064&	0.6596& 0.0702&	0.4526 \\
          Chatglm-9B \citep{glm2024chatglm} & 0.3666&	0.6902& 0.2530&	0.4736&	0.3514&	0.6820&0.2397&	0.5000&	0.1832&	0.6506& 0.1192&	0.4342 \\
          Llama3.2-1B-Instruct \citep{dubey2024llama} & 0.0644&	0.5055& 0.0326&	0.1664&	0.0822&	0.4410& 0.0363&	0.1604&	0.0714&	0.4757& 0.0340&	0.1796\\
          Gemma3-12B \citep{team2025gemma}& 0.4528&	0.694& 0.3142&	0.5912&	 	0.3256&	0.3548&	0.1155& 0.4478&	 	0.1666&	0.0869 & 0.0145 &	0.3322 \\
          Deepseek-R1-distill-Qwen-7B~\citep{guo2025deepseek} & 0.4048 & 0.4416 & 0.1788 & 0.5622 & 0.2896 & 0.4535 & 0.1313 & 0.2034 & 0.0835 & 0.4556 & 0.0380 & 0.6074 \\
          SmileyLlama \citep{cavanagh2026smileyllama} & 0.4520 & 0.4960 & 0.2241 & 0.7860 & 0.4070 & 0.5180 & 0.2111 & 0.7730 & 0.2900 & 0.5170 & 0.1499 & 0.7800 \\
          ChemDFM-v1.5-8B \citep{zhao2025developing}&	0.2998&	0.6613&	0.1983&	0.7582&	0.2510&	0.6831&	0.1715&	0.8170	&0.3112&0.6600	&0.2054	&0.8072\\
          ChemLLM-20B \citep{zhang2024chemllm} & 0.2992&	0.5634&	0.1686&	0.3788	&0.1316& 0.5452&	0.0717&	0.293&	0.2594&	0.5739&	0.1489&	0.3856\\
          \midrule
          MolT5-small \citep{edwards2022translation} & 0.2158&	0.1052& 0.0227&	0.4302&	0.2316&	0.1011& 0.0234&	0.4420&	0.2214&	0.1031& 0.0228&	0.4326 \\
          MolT5-base \citep{edwards2022translation} & 0.2074&	0.1051& 0.0218& 0.4168&	0.1856&	0.1073& 0.0199&	0.3796&	0.2358&	0.1054& 0.0249&	0.4536 \\
          MolT5-large \citep{edwards2022translation} & 0.4244&	0.1015& 0.0431&	0.8156&	0.4496&	0.1072& 0.0482&	0.8678&	0.4654&	0.1190& 0.0554&	0.9214 \\
          BioT5-base \citep{pei2023biot5} & 0.5158&	0.1526& 0.0787&	1.0000&	0.5060&	0.1597& 0.0808&	1.0000&	0.5068&	0.1580& 0.0801&	1.0000 \\
          \midrule
          Llama3.2-1B (OpenMolIns-large)& 0.2898&	0.5951& 0.1725&	0.3850&	0.2644&	0.5956&0.1575&	0.3678&	0.1996&	0.5849&0.1167&	0.3490\\
          Llama3.1-8B (OpenMolIns-large)& \colorbox{backblue!75}{0.8054}&	0.6678&0.5378&	0.8720&	0.7122&	0.6548&0.4663&0.8514&	0.5224&	0.6398& 0.3342&	0.8802 \\
          Llama3.1-8B (OpenMolIns-xlarge)& \colorbox{backred!50}{0.8822}&	0.6662&\colorbox{backred!50}{0.5877}&	0.9314&	0.6982&	0.6693&0.4673&0.9422&	\colorbox{backred!50}{0.8648}&	0.6736& \colorbox{backred!50}{0.5825}&	0.9310 \\
          Galactica-125M (OpenMolIns-light) & 0.3202&	0.6547& 0.2096&	0.6416&	0.3508&	0.6435&	0.2257&0.6358&	0.2690&	0.6521& 0.1754&	0.6380  \\
          Galactica-125M (OpenMolIns-small) & 0.4172&	0.6420& 0.2678&	0.5568&	0.3958&	0.6452&	0.2554&0.5338&	0.2956&	0.6385&0.1887&	0.5376 \\
          Galactica-125M (OpenMolIns-medium) & 0.5904&	0.5812& 0.3431&	0.7890&	0.5874&	0.5873&	0.3450& 0.7384&	0.4608&	0.5859& 0.2700&	0.7768 \\
          Galactica-125M (OpenMolIns-large) & 0.6454&	0.5927& 0.3825&	0.8198&	0.6388&	0.5973&0.3816&	0.8028&	0.4950&	0.5962&0.2951&	0.8100 \\
          Galactica-125M (OpenMolIns-xlarge) & 0.7362&	0.5744&0.4229&	0.8902&	0.7124&	0.5697&	0.4059&0.8612&	\colorbox{backblue!75}{0.5786}&	0.5677& 0.3285&	0.8626 \\
     \bottomrule
     \end{tabular}
     }

     \label{tab:molopt}
 \end{table}

 \begin{table}[t]
     \centering
     \caption{Detailed results for subtasks on MolCustom. For each task, we highlight the \colorbox{backred!50}{best} and the \colorbox{backblue!75}{second-best} success rate (SR), as well as the weighted success rate (WSR).}
     \resizebox{1.0\columnwidth}{!}{
     \begin{tabular}{c|cccc|cccc|cccc}
     \toprule
         \multirow{2}{*}{Models} & \multicolumn{4}{c|}{AtomNum} & \multicolumn{4}{c|}{BondNum} & \multicolumn{4}{c}{FunctionalGroup}\\ \cline{2-13}
          & SR & Novelty & WSR & Validity & SR & Novelty & WSR & Validity & SR & Novelty & WSR & Validity \\
     \midrule
          GPT-4o \citep{achiam2023gpt} & \colorbox{backred!50}{0.1998} & 0.6703 & \colorbox{backred!50}{0.1339} & 0.5852 & 0.0650 & 0.6336 & 0.0412 & 0.8564 & 0.2330 & 0.6513 & 0.1518 & 0.8590 \\
          GPT-4-turbo \citep{achiam2023gpt} & 0.1702 & 0.6991 & 0.1190 &	0.4904&	0.0774&	0.6301& 0.0488&	0.9068&	0.2180&	0.6605& 0.1440&	0.8778 \\
          GPT-3.5-turbo \citep{achiam2023gpt} & 0.1070&	0.5054& 0.0541 &	0.6947&	0.0518&	0.6871& 0.0356 &	0.5522&	0.1136&	0.6585& 0.0748&	0.8686 \\
          Claude-3.5 \citep{claude3.5} & \colorbox{backblue!75}{0.1928} &	0.6926& \colorbox{backblue!75}{0.1335}&	0.6548&	0.1058 &	0.6584& 0.0697&	0.8860&	0.2364&	0.6582&0.1556&	0.8892 \\
          Claude-3 \citep{claude3} & 0.1044&	0.6833& 0.0713 &	0.5910&	0.1042&	0.6598& 0.0688 &	0.8696&	0.1816&	0.9158& \colorbox{backblue!75}{0.1663} &	0.6644 \\
          Gemini-1.5-pro \citep{gemini} & 0.1742&	0.6902& 0.1202 &	0.6774&	0.0708&	0.6522& 0.0462 &	0.8688&	\colorbox{backblue!75}{0.2486}&	0.6673& 0.1659 &	0.9240 \\
          \midrule
          Llama3-70B-Instruct (Int4) \citep{dubey2024llama} & 0.1404&	0.6675& 0.0937	&0.5474&	0.0670&	0.6478& 0.0434 &	0.7378&	0.1752&	0.6576& 0.1152 &	0.7650 \\
          Llama3-8B-Instruct \citep{dubey2024llama} & 0.0242&	0.6649&	0.0161 &0.3812&	0.0260&	0.6303&	0.0164 &0.5700&	0.0848&	0.6167& 0.0523 &	0.7216 \\
          Llama3.1-8B-Instruct \citep{dubey2024llama} & 0.0228 & 0.7020 & 0.0160 & 0.3862 & 0.0395 & 0.6541 & 0.0258 & 0.6387 & 0.1300 &0.6274 &0.0816 & 0.6905 \\
          Mistral-7B-Instruct-v0.2 \citep{jiang2023mistral}& 0.0078&	0.6732& 0.0053&	0.2986&	0.0102&	0.6309& 0.0064&	0.4524&	0.0048&	0.6012& 0.0029&	0.4020 \\
          Qwen2-7B-Instruct \citep{yang2024qwen2} & 0.0110&	0.9061& 0.0100 &0.2622&	0.0010&	0.8645& 0.0090 &	0.0796&	0.0022&	0.8601& 0.0019&	0.0622 \\
          Yi-1.5-9B \citep{young2024yi} & 0.0392&	0.6848& 0.0268 &	0.6170&	0.0208&	0.6407& 0.0133&	0.7072&	0.0126&	0.6945& 0.0088&	0.6521 \\
          Chatglm-9B \citep{glm2024chatglm} & 0.0002&	0.7483& 0.0001&	0.2131&	0.0254&	0.7189& 0.0183&	0.4682&	0.0000&	0.6908& 0.0000 &	0.5926 \\
          Llama3.2-1B-Instruct \citep{dubey2024llama} & 0.0040&	0.6807& 0.0027 &0.1850&	0.0080&	0.7465& 0.0060 &	0.2226&	0.0008&	0.7461& 0.0006&	0.2818\\
          Gemma3-12B \citep{team2025gemma} & 0.0460&	0.6017&	0.0277& 0.3414&	 	0.0280&	0.5973&	0.0167 &	0.3318& 	0.0006& 0.5940&	0.0004 &	0.3224 \\
          Deepseek-R1-distill-Qwen-7B~\citep{guo2025deepseek} & 0.1574 & 0.6611 & 0.1041 & 0.4815 & \colorbox{backblue!75}{0.1631} & 0.7264 & \colorbox{backred!50}{0.1185} & 0.5758 & 0.0385 & 0.6248 & 0.0241  & 0.5634 \\
          SmileyLlama \citep{cavanagh2026smileyllama} & 0.0002 & 0.6560 & 0.0001 & 0.6420 & 0.0116 & 0.5970 & 0.0069 & 0.6990 & 0.0052 & 0.5970 & 0.0031 & 0.7430 \\
          ChemDFM-v1.5-8B \citep{zhao2025developing}&	0.0112&	0.6809&	0.0076&	0.6700&	0.0172&	0.6146&	0.0106&	0.8968&	0.0312&	0.6011&	0.0188&	0.8094\\
          ChemLLM-20B \citep{zhang2024chemllm} & 0.0288&	0.7055&	0.0203&	0.3530&	0.0204&	0.6786&	0.0138&	0.3678&	0.0600&	0.6452&	0.0387&	0.5442\\
          \midrule
          MolT5-small \citep{edwards2022translation} & 0.0006&	0.6586& 0.0004&	0.6610&	0.0064&	0.5980& 0.0038 &	0.6202&	0.0114&	0.5287& 0.0060&	0.8354 \\
          MolT5-base \citep{edwards2022translation} & 0.0008&	0.6868& 0.0005&	0.7560&	0.0070&	0.6509& 0.0046 &	0.8422&	0.0130&	0.5464& 0.0071&	0.8382 \\
          MolT5-large \citep{edwards2022translation} & 0.0150&	0.7103& 0.0107&	0.8412&	0.0118&	0.5611& 0.0066 &	0.8916&	0.0382&	0.6088& 0.0233&	0.9406\\
          BioT5-base \citep{pei2023biot5} & 0.0118&	0.8353& 0.0099&	0.9950&	0.0078&	0.6667& 0.0052&	0.9992&	0.0476&	0.6792& 0.0323&	0.9998 \\
          \midrule
          Llama3.2-1B (OpenMolIns-large)& 0.0144&	0.6490& 0.0093 &	0.5616&	0.0350&	0.6150& 0.0215 &	0.6186&	0.0252&	0.6373& 0.0161 &	0.4412\\
          Llama3.1-8B (OpenMolIns-large)& 0.0136&	0.6634& 0.0090&	0.7582&	0.0544&	0.6614& 0.0360&	0.7456&	0.1344&	0.6396& 0.0860&	0.6435 \\
          Llama3.1-8B (OpenMolIns-xlarge)& 0.1186&	0.6834& 0.0811&	0.8540&	0.1196&	0.6746& 0.0807&	0.9000&	\colorbox{backred!50}{0.3548}&	0.6393& \colorbox{backred!50}{0.2268}&	0.9492 \\
          Galactica-125M (OpenMolIns-light) &0.0044	&0.6054& 0.0026&	0.7930&	0.0216&	0.5724& 0.0124&	0.7596&	0.0244&	0.5756& 0.0140&	0.8442 \\
          Galactica-125M (OpenMolIns-small) & 0.0146&	0.6568&	0.0096&0.8424&	0.0530&	0.6365& 0.0337&	0.7926&	0.0570&	0.5954& 0.0339&	0.8874 \\
          Galactica-125M (OpenMolIns-medium) & 0.0294&	0.6553& 0.0193&	0.8698&	0.0622&	0.6473& 0.0403&	0.7474&	0.0882&	0.6091& 0.0537&	0.8932 \\
          Galactica-125M (OpenMolIns-large) & 0.0464&	0.6729&	0.0312&0.9116&	0.0716&	0.6695& 0.0479&	0.7374&	0.0996	&0.6276& 0.0625&	0.8966 \\
          Galactica-125M (OpenMolIns-xlarge) & 0.1862&	0.6899&	0.1285 &0.9308&	\colorbox{backred!50}{0.1656}&	0.6887& \colorbox{backblue!75}{0.1140}&	0.7952&	0.2006&	0.6445& 0.1293&	0.9162 \\
     \bottomrule
     \end{tabular}
     }
     
     \label{tab:molcustom}
 \end{table}

\section{Limitations}
\label{sec:lim}
Although S\textsuperscript{2}-Bench is carefully designed and well-validated through our experiments, we still observe several limitations:

\noindent\textbf{Data Distribution.} Our allocation of atoms, bonds, and functional groups is largely empirical and may not be sufficiently accurate to reconstruct real-world scenarios. 

\noindent\textbf{Application to Complex Real-world Constraints.} Incorporating constraints like stereochemistry, 3D geometry, ring/bridge topology, and basic synthesizability would represent more important aspects of real-world molecular design.

\noindent\textbf{Realism of Instructions.} Template-generated prompts may not fully capture the richness of real-world chemist queries.

\section{Broader Impacts}
\label{sec:impact}
S\textsuperscript{2}-Bench firstly proposes the text-based open molecule generation task for LLMs, aiming at adopting LLMs as chemist assistants and molecule operators for molecule discovery. Hopefully, our benchmark could inspire more research in this promising area, thereby accelerating the pace of drug discovery and material design. Ultimately, the development in molecule discovery will benefit human welfare in various ways. For example, the effective medicines could help save people's lives and expand their lifespan.

\section{GenAI Disclosure}
During the preparation of this work, the author(s) used LLMs to improve the language and readability. After using this tool/service, the author(s) reviewed and edited the content as needed and take(s) full responsibility for the content of the publication.

\clearpage

\end{document}